\documentclass[11pt]{article}

\usepackage[margin=1in]{geometry}
\usepackage{amsmath, amssymb}
\usepackage{hyperref}
\usepackage{enumitem}
\usepackage{xcolor}
\usepackage{xr}          
\usepackage{hyperref}    
\usepackage{xr-hyper}    
\usepackage{graphicx}
\usepackage{booktabs} 
\usepackage{soul}

\usepackage{algorithm}
\usepackage{algpseudocode}
\usepackage{tikz}
\usepackage{times}
\usepackage{latexsym}
\usepackage{amsthm}
\usepackage{amsmath}
\usepackage{amssymb}
\usepackage{multirow}
\usepackage{graphicx}
\usepackage{threeparttable}
\usepackage{algpseudocode}
\usepackage{color,soul}
\usepackage{booktabs}

\usepackage{caption}

\usepackage[labelfont=sf]{subcaption}

\usepackage[most]{tcolorbox}
\usepackage[utf8]{inputenc}
\usepackage{enumitem}

\newcommand{\ABSTRACT}[1]{%
  \begin{abstract}
  #1
  \end{abstract}
}

\newcommand{\KEYWORDS}[1]{%
  \vspace{0.5em}
  \noindent\textbf{Keywords:} #1
}

\usepackage[numbers]{natbib}

\begin{document}

\title{A Role-Based LLM Framework for Structured Information Extraction from Healthy Food Policies}




\author{
  Congjing Zhang\textsuperscript{a}\thanks{Equal contribution.} \quad
  Ruoxuan Bao\textsuperscript{b}\footnotemark[1] \quad
  Jingyu Li\textsuperscript{c} \quad
  Yoav Ackerman\textsuperscript{a} \quad
  Shuai Huang\textsuperscript{a}\thanks{Corresponding author.} \\
  Yanfang Su\textsuperscript{d}\footnotemark[2]
  \\[8pt]
  \textsuperscript{a}Department of Industrial \& Systems Engineering, University of Washington, Seattle, WA, \\USA;
  \textsuperscript{b}Department of Management, Shanghai University, Shanghai, China;
    \textsuperscript{c}H. Milton Stewart \\School of Industrial and Systems Engineering, Georgia Institute of Technology, Atlanta, GA,\\ USA;
      \textsuperscript{d}Department of Global Health, University of Washington, Seattle, WA, USA\\
  [6pt]
    \texttt{\{congjing, yoava, shuaih, yfsu\}@uw.edu} \\ \texttt{chris20030909@gmail.com} \quad 
  \texttt{jli928@gatech.edu} 
}

\date{}

\maketitle

\ABSTRACT{%
Current Large Language Model (LLM) approaches for information extraction (IE) in the healthy food policy domain are often hindered by various factors, including misinformation, specifically hallucinations, misclassifications, and omissions that result from the structural diversity and inconsistency of policy documents. To address these limitations, this study proposes a role-based LLM framework that automates the IE from unstructured policy data by assigning specialized roles: an LLM policy analyst for metadata and mechanism classification, an LLM legal strategy specialist for identifying complex legal approaches, and an LLM food system expert for categorizing food system stages. This framework mimics expert analysis workflows by incorporating structured domain knowledge, including explicit definitions of legal mechanisms and classification criteria, into role-specific prompts. We evaluate the framework using 608 healthy food policies from the Healthy Food Policy Project (HFPP) database, comparing its performance against zero-shot, few-shot, and chain-of-thought (CoT) baselines using Llama-3.3-70B. Our proposed framework demonstrates superior performance in complex reasoning tasks, offering a reliable and transparent methodology for automating IE from health policies.
}%






\section{Introduction}
Public health policymaking in the United States operates at an unprecedented scale and complexity \cite{brownson2009understanding}. Each year, thousands of health-related policies are enacted across federal, state, and local jurisdictions \cite{bruzelius2024punitive}. These policies represent critical interventions that shape population health outcomes, yet their sheer volume and diversity create fundamental challenges for researchers seeking to understand how they are implemented \cite{walt2008doing}. The emergence of Large Language Models (LLMs) offers a promising opportunity to address this scalability challenge. Their ability to process natural language without task-specific training makes them particularly attractive for policy research, where heterogeneous document formats and rapidly evolving domains limit the feasibility of building specialized extraction systems \cite{harris2024evaluating}.

However, directly applying LLMs to health policy IE remains challenging and underexplored. Unlike standardized datasets with consistent schemas, health policy data exhibits characteristics that make it highly prone to LLM misinformation \cite{milanese2025fact}. First, policies span multiple legal mechanisms (e.g., laws, executive orders, administrative rules) and substantive categories, requiring domain-specific reasoning to classify correctly. Second, data is collected from heterogeneous sources with inconsistent formats, terminology, and levels of detail, often necessitating tailored extraction pipelines. Third, accurate categorization requires both parsing formal legal structures and interpreting policy intent—tasks that exceed the capabilities of generic prompting. These challenges manifest as hallucinations, misclassifications, and omissions, which can undermine downstream policy analysis and decision-making \cite{ji2023towards, mustafa2025large, cai2024assessment}.

Existing approaches to improving LLM-based extraction typically lie at two extremes. On one end, heavyweight solutions such as fine-tuning or agent-based systems can improve performance but require substantial labeled data, engineering overhead, and system complexity, limiting their practicality in rapidly evolving policy domains. On the other end, lightweight methods such as zero-shot, few-shot, or chain-of-thought (CoT) prompting are easy to deploy but often lack structured reasoning and fail to handle multi-dimensional classification tasks reliably.

In this work, we position our approach as a middle ground: a lightweight yet structured LLM-based alternative that improves information extraction (IE) reliability without incurring the cost of full system redesign. We propose a role-based prompting framework that decomposes policy extraction into specialized analytical roles. Specifically, the framework assigns the LLM three complementary roles: a Policy Analyst responsible for metadata and mechanism classification, a Legal Strategy Specialist for identifying complex legal approaches, and a Food System Expert for categorizing stages within the food system. To support domain-aware reasoning, we incorporate structured domain knowledge, such as policy categorization schemes, legal mechanisms, and classification criteria, directly into role-specific prompts. This design introduces modular task decomposition within a purely prompt-based architecture, enabling more consistent and interpretable reasoning while preserving the flexibility and low operational cost of prompting-based methods. By bridging the gap between simple prompting strategies and resource-intensive system redesigns, our framework offers an efficient and practical solution for scalable health policy IE.





\begin{figure}[t]
  \centering
  \includegraphics[width=0.8\linewidth]{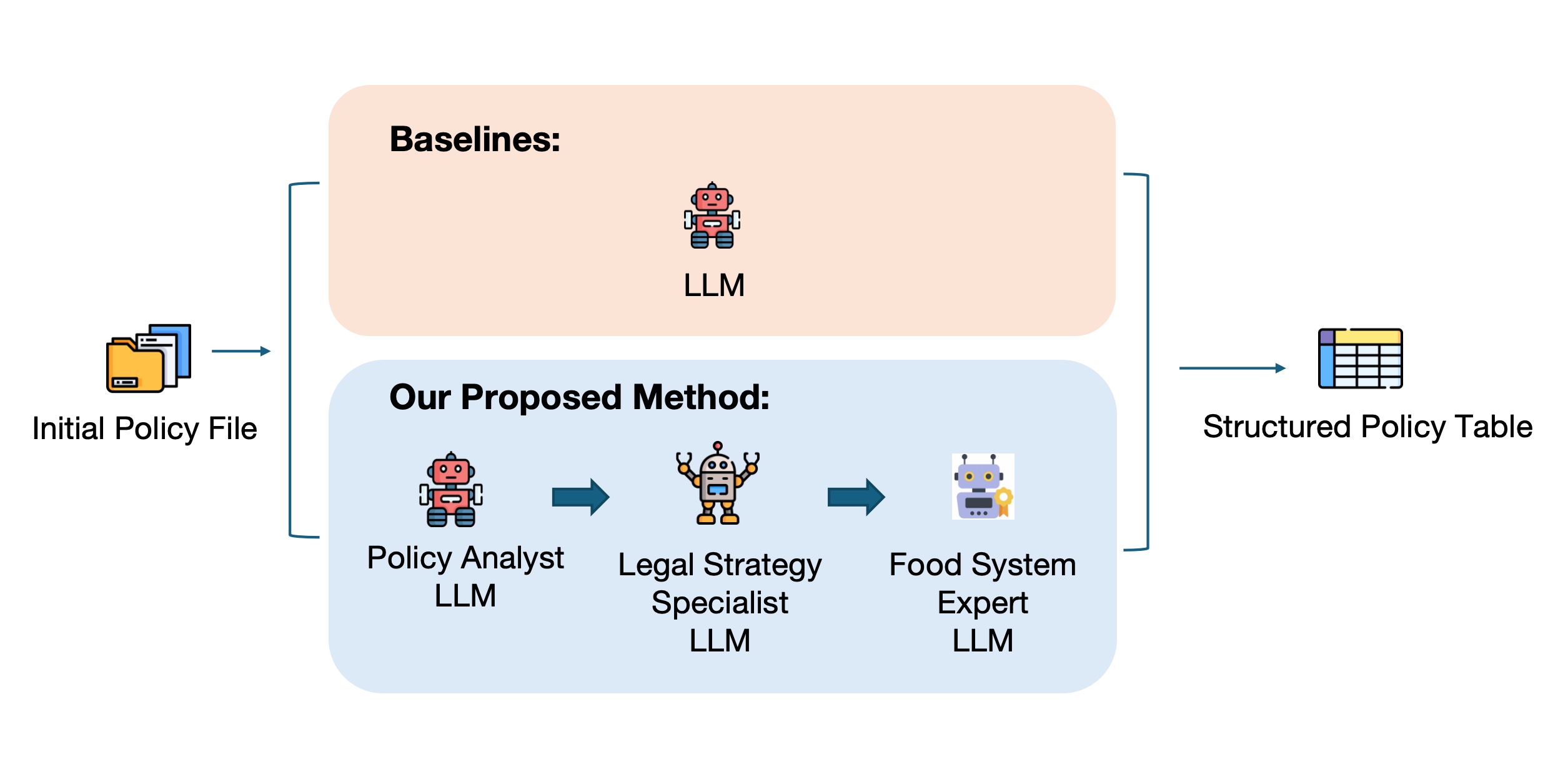}
  \caption{The framework of our method to extract structured information from healthy food policies.}
  \label{framework}
\end{figure}

\section{Literature Review}
\subsection{Information Extraction via LLMs}
IE has long been a fundamental task in natural language processing (NLP), aiming to convert unstructured text into structured, machine-readable representations. Traditional approaches, such as rule-based systems and statistical models, often perform well in narrowly defined settings, achieving high precision where domain rules and features are carefully crafted. However, their reliance on manual engineering and task-specific heuristics severely limits scalability across longer texts, diverse domains, or evolving language patterns.

Recent advances in LLMs have significantly reshaped the IE landscape. Pretrained on massive and diverse corpora, LLMs possess extensive knowledge across various domains and strong generalization capabilities. This pretrained knowledge enables them to perform extraction tasks with minimal supervision, often in zero-shot or few-shot settings, and without additional training. As a result, LLM-based IE methods have demonstrated competitive or even state-of-the-art performance across a variety of domains, including legal documents, and web content \cite{Fornasiere_Brunello_Scotti_Carman}. One line of research adopts hybrid pipelines, in which rule-based systems are first used to generate candidate extractions, followed by LLMs that validate, refine, or complete these outputs. Such designs retain the precision and interpretability of symbolic rules while leveraging the contextual reasoning ability of LLMs, leading to improved robustness in complex or noisy documents \cite{dagli2024development}. In parallel, fine-tuned small language models have been shown to achieve strong extraction performance when adapted to task-specific schemas, offering a more computationally efficient alternative to large-scale LLMs \cite{peng2024metaie}. These approaches highlight that effective IE does not depend on model size alone, but rather on aligning model capacity with the structure and complexity of the extraction task. Prompt-based extraction with LLMs has further expanded the methodological landscape by enabling zero-shot and few-shot IE without explicit supervision. However, prompting-based approaches are often sensitive to prompt design and input variability. To address these limitations, retrieval-augmented strategies have been introduced, incorporating relevant documents or schema descriptions into the prompt context to reduce hallucination \cite{liu2023improving}. Together, these lines of work suggest a growing trend toward more structured IE frameworks integrating LLM reasoning ability.

\subsection{AI for Public Health and Policy Analysis}
As the volume and heterogeneity of data continue to expand across domains from healthcare and public health to legal compliance and policy decision-making, traditional policy analysis techniques face increasing challenges. However, with the introduction of AI, there holds a possibility in which efficiency and accuracy can be empowered. Not only that, but AI could also improve population health outcomes \cite{panahi2025ai}. AI has the power to tackle large quantities of data in a fraction of the time it would normally take; this can further lead to possible early detection of diseases. This was especially evident during the COVID-19 pandemic, when companies such as BlueDot and MetaBiota utilized AI-driven algorithms to detect the outbreak in China before it was detected by the rest of the world \cite{allam2020artificial}. 

AI can be utilized to monitor how we tackle policy changes; with that, we can determine how impactful or detrimental they are on various populations \cite{ramezani2023application}. It is true that how we determine health policy is heavily reliant on data. This data can be from various areas such as population, clinical, and epidemiology. These become the backbone for decision-making for our health policies \cite{panahi2025ai}. In current states of record-based policymaking, it takes more resources as well as computing power to analyze and determine outcomes. AI provides the capability to analyze datasets rapidly, which in turn can lead to more efficient and effective policies. Not only that, but with the rapid development of larger datasets and AI being used to analyze them, we open a new window of a larger impact onto predictive modeling for policy making. Utilizing its ability to model various policy options, policymakers can make more diverse and better-informed decisions \cite{panahi2025ai}.


With every tool we use, we have to understand that there are limits to its usage. We break down the idea of policy analysis into two key pillars; these are known as data collection and data analysis \cite{shi2017can}. As mentioned, AI is able to handle large amounts of data and provide much more rapid and effective results in terms of policy analysis. However, with all that power, we run into a point being that misusage of the AI can cause it to not properly understand the real world. There needs to be a government-provided policy that provides more control over evidence building \cite{shi2017can}.

\begin{figure}[t]
  \centering
  \includegraphics[width=0.9\linewidth]{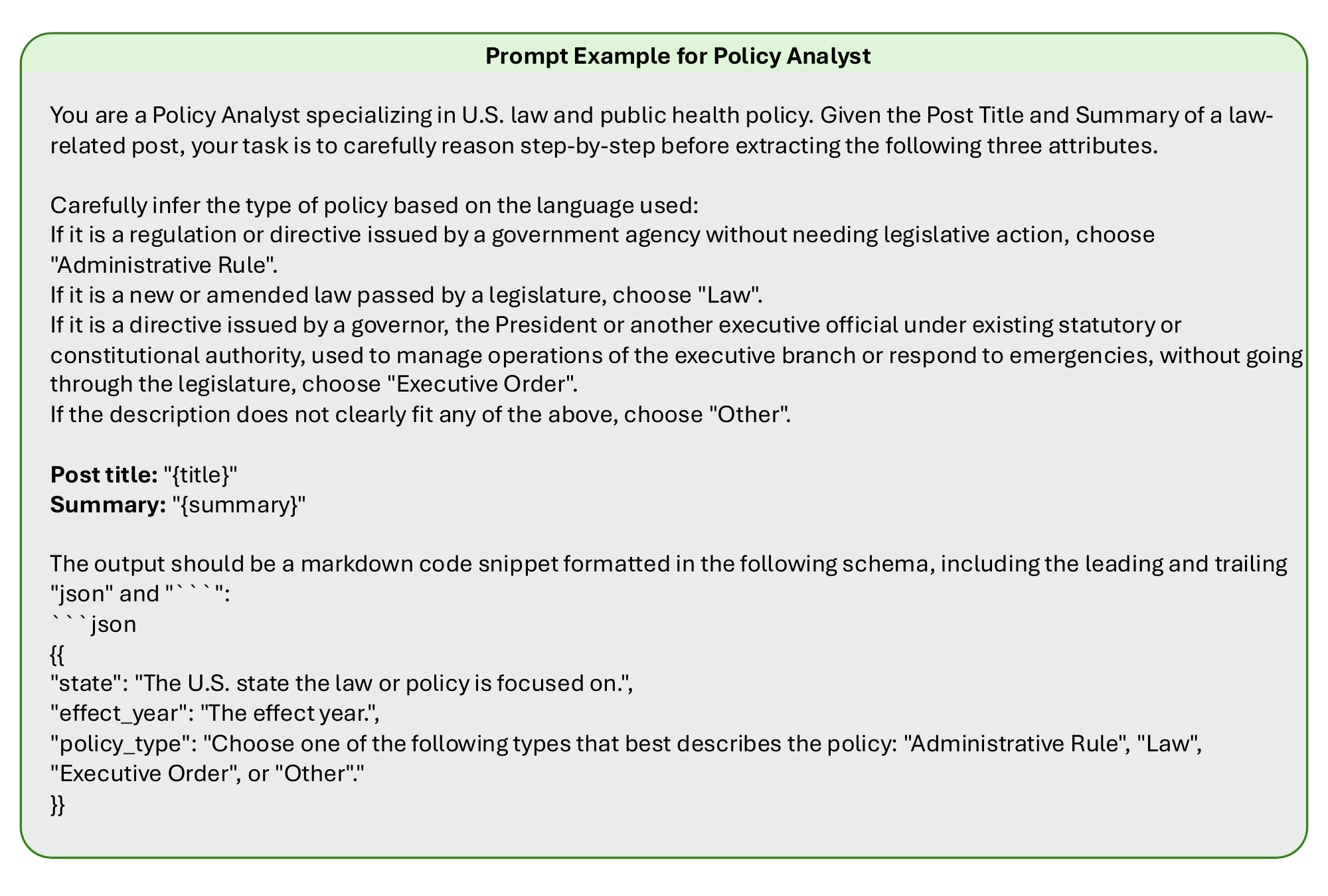}
  \caption{Prompt for LLM Policy Analyst.}\label{policy_analyst}
\end{figure}

While AI can have the cons addressed, it is through AI we have also been able to make leaps in imaging. With its data processing capabilities we can see how it's able to take such large data and then produces images that can be used to develop diagnostic tools, which in turn helps drive an increase in the speed and accuracy with which diseases can be detected as well as lighten the burden healthcare systems take \cite{chumachenko2025artificial}. IE has become a very valuable asset for health policy, as with all the data we have collected AI now has the power to take all that data and perform structured analysis that provides governments as well as policymakers the opportunity to better prepare and allocate the proper resources \cite{chumachenko2025artificial}.

\begin{figure}[t]
  \centering
  \includegraphics[width=0.9\linewidth]{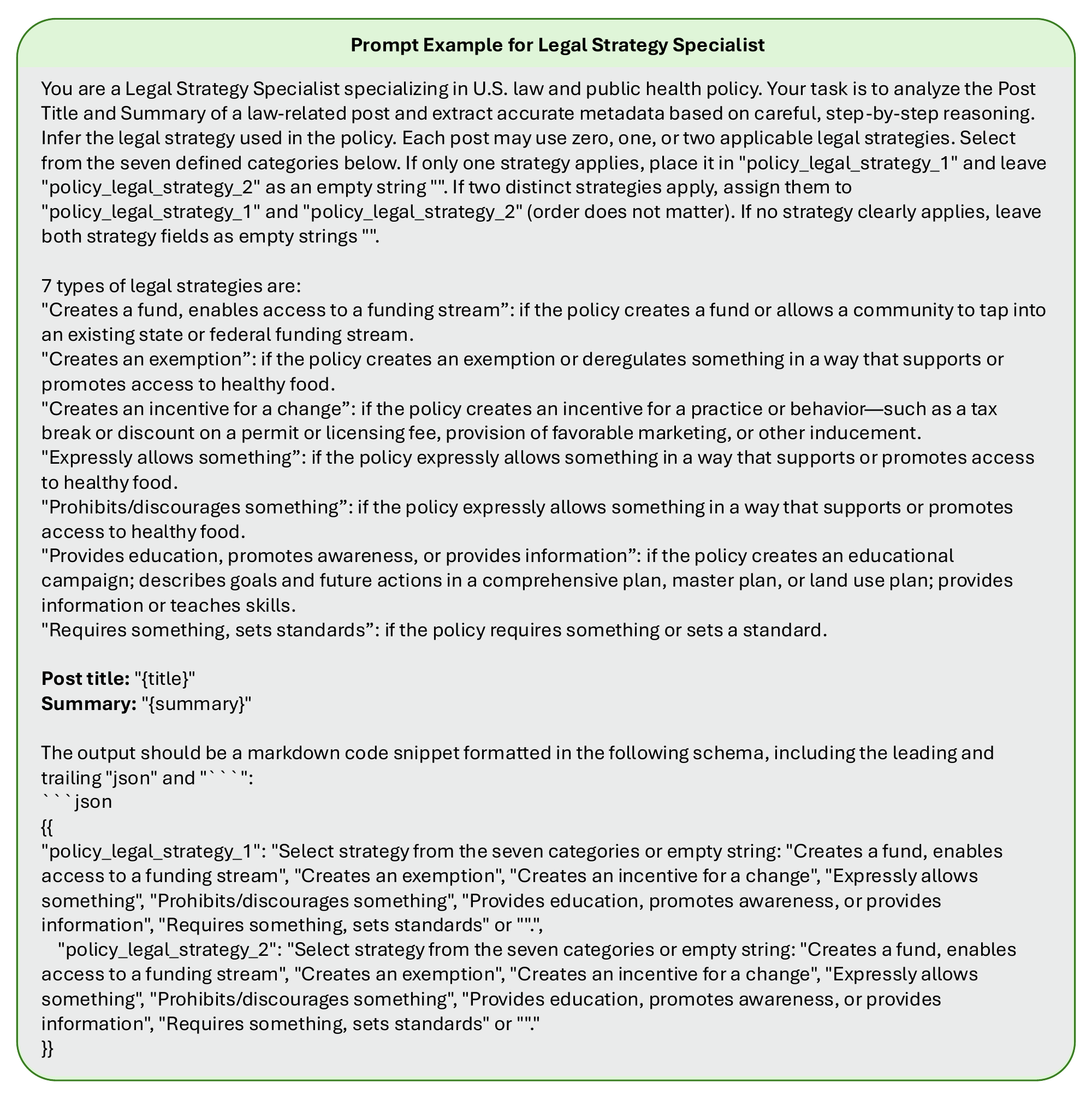}
  \caption{Prompt for LLM Legal Strategy Specialist.}\label{legal_strategy}
\end{figure}

\section{Methods}
We propose a role-based LLM framework that assigns specialized roles to multiple agents, each configured with domain-specific expertise. Figure \ref{framework} provides an overview of our framework. By utilizing carefully crafted prompts that incorporate structured knowledge of policy classification and legal mechanisms, the framework ensures that the extracted information captures the full regulatory lifecycle and thematic scope of health interventions. The framework systematically processes policy titles and summaries to extract five key attributes, chosen to provide a comprehensive profile of each entry:
\begin{itemize}
    \item Implementation State \& Effective Year: to track the geospatial and temporal distribution of policy adoption.
    \item Policy Type \& Legal Strategies: to categorize the regulatory nature and enforcement mechanisms used to drive public health outcomes.
    \item Food System Categories: to map the specific intervention points within the food supply chain.
\end{itemize}
By employing three specialized LLM roles, each tailored to a distinct extraction task, the framework maintains high precision across these varied dimensions, transforming raw text into a structured, multi-dimensional dataset.

Each role-specific prompt follows a structured format: (1) The prompt provides clear specification of the LLM's expertise and analytical perspective (e.g., ``You are a Policy Analyst specializing in U.S. law and public health policy") based on the extraction task. (2) The explicit instructions for the extraction task are given to the LLM, including requirements for step-by-step reasoning before generating outputs. (3) The prompt includes comprehensive definitions of classification categories, decision criteria, and edge case handling to reduce misinformation. (4) Both policy title and summary are provided as inputs. (5) Structured JSON format specification are included to ensure consistent and machine-readable outputs. The prompts are designed to leverage LLMs' ability to simulate domain expertise while providing explicit guidance that standardizes the extraction process and reduces the likelihood of misinformation.

\begin{figure}[htbp]
  \centering
  \includegraphics[width=0.9\linewidth]{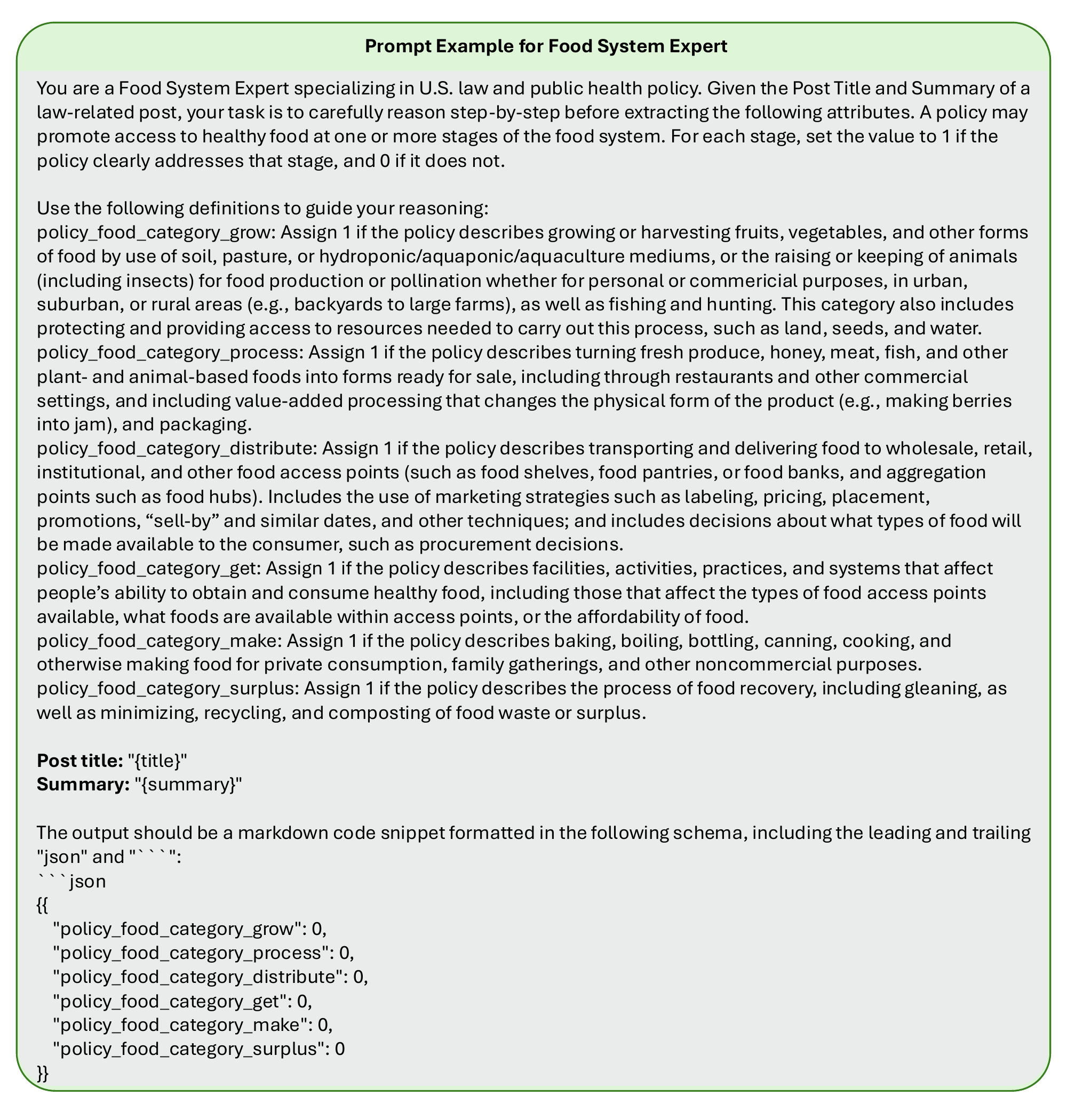}
  \caption{Prompt for LLM Food System Expert.}
  \label{food_system}
\end{figure}

\paragraph{LLM Policy Analyst.} This role specializes in extracting fundamental policy metadata, including the state of implementation, effective year, and policy type classification. Figure \ref{policy_analyst} provides the prompt for the LLM Policy Analyst. The Policy Analyst distinguishes between four policy types based on legal mechanisms: Administrative Rules (regulations or directives issued by government agencies), Laws (legislation passed by legislatures), Executive Orders (directives issued by executives under existing authority), and Other (policies not clearly fitting the previous categories). The prompt provides explicit definitions and decision criteria for each category to reduce ambiguity in classification.

\paragraph{LLM Legal Strategy Specialist.} This role focuses on identifying the legal strategies employed by policies. Figure \ref{legal_strategy} shows the prompt for the LLM Legal Strategy Specialist. The specialist is tasked with recognizing up to two legal strategies per policy from seven predefined categories: (1) Creates a fund or enables access to funding streams, (2) Creates an exemption, (3) Creates an incentive for change, (4) Expressly allows something, (5) Prohibits or discourages something, (6) Provides education, promotes awareness, or provides information, and (7) Requires something or sets standards. The prompt explicitly instructs the model to assign strategies only when clearly applicable, leaving fields empty when no strategy is evident, thereby reducing hallucination risks.

\paragraph{LLM Food System Expert.} This role analyzes how policies promote access to healthy food across different stages of the food system. Figure \ref{food_system} shows the prompt for the LLM Food System Expert. The expert assigns binary indicators (0 or 1) for six food system categories: Grow (food production and resource access), Process (transforming fresh foods for sale), Distribute (transportation and marketing), Get (facilities and systems affecting food access), Make (non-commercial food preparation), and Surplus (food recovery and waste management). Each category is accompanied by detailed definitions in the prompt to ensure consistent interpretation across diverse policy contexts.

\section{Experiments and Results}
\subsection{Experiment Settings}
\paragraph{Databases}
To evaluate our framework, we utilize the Healthy Food Policy Project (HFPP) database \cite{hfpp}, which contains 608 healthy food policies formally adopted by municipal governments across 51 U.S. states and territories. These policies were compiled from diverse sources, including the Growing Connections policy database, web searches, and municipal legal code libraries such as American Legal Publishing and Municode. Each entry includes comprehensive metadata for one specific policy. We derive the ground truth labels used for evaluation from the original HFPP expert-annotated classifications and treat them as the gold standard for our multi-label learning task.

\paragraph{Evaluation Metrics.} 
Rather than traditional machine learning baselines (e.g., SVM or Random Forest), this decision is driven by the complex, multi-faceted nature of policy language, where LLMs offer superior zero-shot reasoning and semantic understanding without the extensive feature engineering required by traditional models. To maintain rigorous evaluation, we provide explicit prompt examples to guide the LLM's classification; however, we apply a strict accuracy constraint where any output category not present in our predefined label set is treated as an inaccuracy (hallucination). Performance is assessed using standard multi-label learning metrics, specifically Micro-F1 and Hamming Loss, which effectively account for the label density and class imbalances inherent in legal policy datasets.


\paragraph{LLM Implementation.} 
We employ Llama-3.3-70B-Instruct-Turbo as the primary LLM backbone via Together AI API for the experiments of all methods in an offline environment. To further assess the generalizability of our proposed framework across different model backbones, we additionally run Qwen-3-80B for our method while keeping the overall framework unchanged. This allows us to isolate the contribution of the role-based design from the underlying language model capacity. To ensure deterministic and reproducible outputs, we set the decoding temperature to $0$ for both our proposed framework and all baseline models.

\paragraph{Baselines.}
To evaluate the effectiveness of our role-based framework, we compare it against three standardized prompting strategies for policy IE: (1) Zero-shot prompting, which involves a direct task description without auxiliary guidance \cite{liu2023pre}; (2) CoT prompting, which provides detailed instructions to elicit step-by-step reasoning \cite{wang2022self}; and (3) Few-shot learning, which extends CoT by providing correctly labeled examples to guide the model's extraction logic \cite{wei2022chain}. For few-shot prompting, we select representative examples from the training set to maximize coverage of policy types and legal mechanisms, while ensuring no overlap with the evaluation set. Unlike our proposed framework, which decomposes complex extraction tasks into specialized roles with domain-specific prompts, all baseline methods are required to extract the complete set of policy attributes within a single, unified prompt. To ensure reproducibility, the specific prompt templates utilized for the baseline models are detailed below. In these templates, $\{title\}$ and $\{summary\}$ serve as placeholders for the input policy metadata.

\begin{tcolorbox}[colback=gray!5!white,colframe=gray!50!black,title=Baseline 1: Zero-Shot Prompting, fonttitle=\bfseries\small, fontupper=\small, arc=1mm]
Given the Post Title and Summary of a law-related post, your task is to complete the following attributes: \texttt{state}, \texttt{effect\_year}, and \texttt{policy\_type} (Choose from: ``Administrative Rule", ``Law", ``Executive Order", or ``Other").\\
\textbf{Post title:} ``\{title\}" \\
\textbf{Summary:} ``\{summary\}" \\
Output the result as a markdown JSON snippet.
\end{tcolorbox}

\begin{tcolorbox}[colback=blue!2!white,colframe=blue!50!black,title=Baseline 2: Chain-of-Thought (CoT) Prompting, fonttitle=\bfseries\small, fontupper=\small, arc=1mm]
Identify the legal attributes step-by-step:
1. \textbf{State Identification}: Scan the title (e.g., ``N.Y.", ``Wash.") to determine the U.S. state.
2. \textbf{Year Extraction}: Locate the specific enactment or revision year within the citations.
3. \textbf{Policy Classification}: Analyze keywords (e.g., ``Code" $\rightarrow$ Administrative Rule) to select the policy type. \\
\textbf{Post title:} ``\{title\}" \\
\textbf{Summary:} ``\{summary\}" \\
Reason first, then output the JSON.
\end{tcolorbox}

\begin{tcolorbox}[colback=green!2!white,colframe=green!50!black,title=Baseline 3: Few-Shot Prompting, fonttitle=\bfseries\small, fontupper=\small, arc=1mm]
Extract attributes based on the following examples:\\
\textbf{Example 1 (Administrative Rule):} Title: ``New York, N.Y., Code..." $\rightarrow$ \{``state": ``New York", ``effect\_year": ``2017", ``policy\_type": ``Administrative Rule"\}\\
\textbf{Example 2 (Law):} Title: ``Seattle, Wash. Ord..." $\rightarrow$ \{``state": ``Washington", ``effect\_year": ``2008", ``policy\_type": ``Law"\}\\
\textbf{Task:} Complete the attributes for Title: ``\{title\}" and Summary: ``\{summary\}".
\end{tcolorbox}

\subsection{Results}
We evaluate the proposed framework on two core tasks: structured IE from health policies and multi-label classification of food system categories. The results collectively demonstrate the effectiveness of role-based decomposition in both precise attribute extraction and holistic policy understanding.

\begin{figure}[h]
  \centering
  \includegraphics[width=0.7\linewidth]{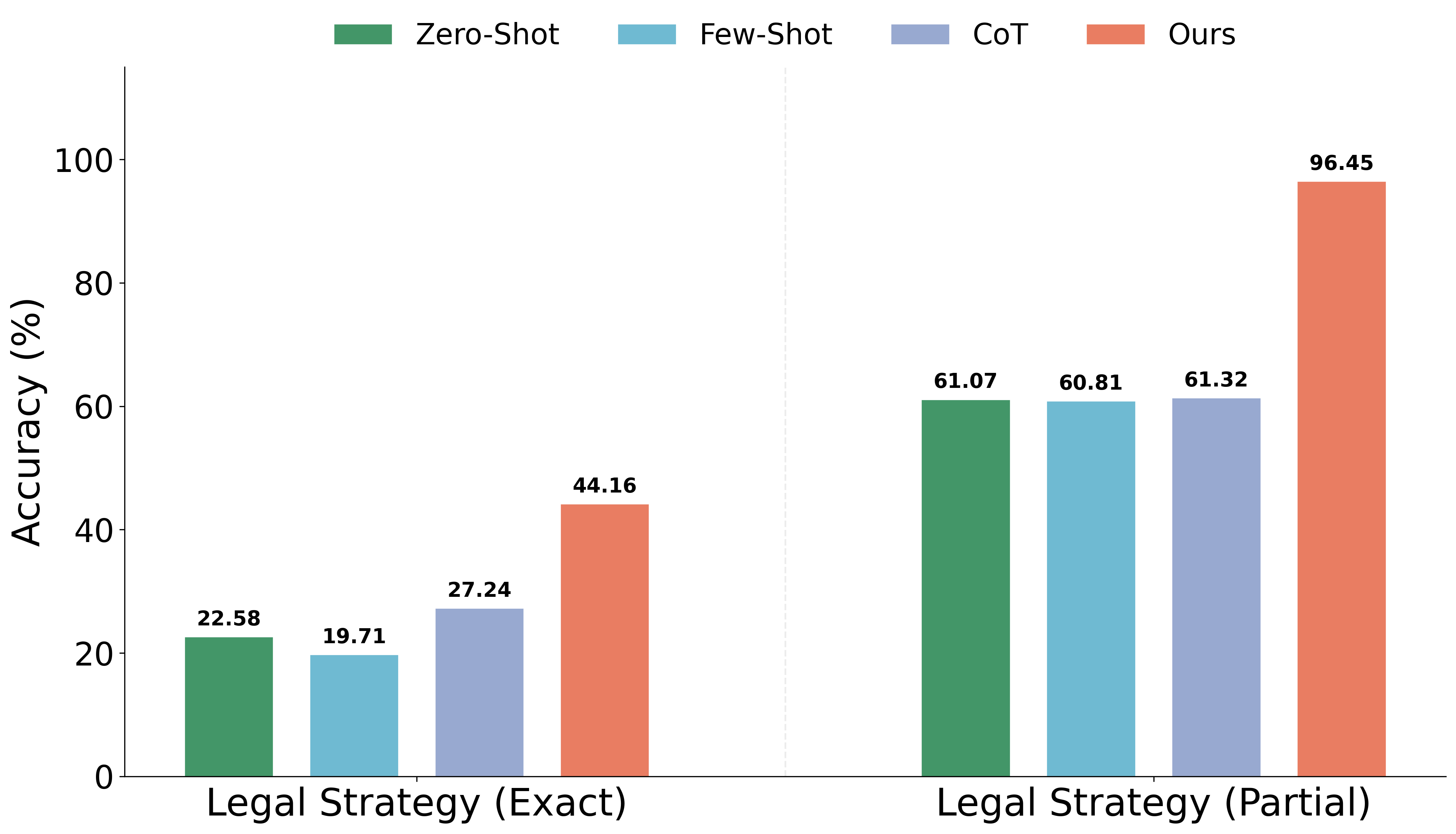}
  \caption{Accuracy of legal strategy for different prompting from Llama-3.3-70B-Instruct-Turbo.}
  \label{visual_leagl}
\end{figure}

Table \ref{tab:health_acc} presents the accuracy results of IE for different methods. For health policy attributes, factual fields such as implementation state and effective year achieve near-perfect performance across all methods, with accuracies consistently above 98\%. This suggests that such information is explicitly stated in policy texts and can be reliably extracted even with simple prompting strategies. However, more complex semantic fields reveal clear differences between methods. In the classification of legal strategies, which requires identifying multiple overlapping mechanisms within a single policy, all baseline methods exhibit relatively low exact-match accuracy, as visualized in Figure \ref{visual_leagl}. Zero-Shot achieves 22.58\%, Few-Shot further drops to 19.71\%, while CoT improves to 27.24\%. In contrast, our framework significantly outperforms all baselines with an exact-match accuracy of 44.16\%, representing a substantial improvement in capturing the full set of applicable strategies. This advantage becomes even more evident under partial-match evaluation, where our method reaches 96.45\%, far exceeding the approximately 61\% achieved by Zero-Shot and CoT. Further analysis at the individual strategy level shows consistent gains across both categories. For Type A strategies (Funding/Access), our framework achieves 43.10\%, outperforming all baselines. For Type B strategies (Incentives/Permissions), Qwen-3 attains the highest score (54.11\%), but Llama-3.3 remains competitive at 45.22\%, indicating balanced performance across different types of policy mechanisms. These results suggest that the proposed role-based design improves both completeness and granularity in multi-label extraction tasks.

\begin{table*}[htbp]
  \centering
  \caption{IE accuracy (\%) for health policy attributes.}
  \small 
  \begin{tabular}{lccccccc}
    \toprule
    \textbf{Methods} & \textbf{State} & \textbf{Year} & \textbf{Policy Type} & \multicolumn{2}{c}{\textbf{Legal Strategy}} & \multicolumn{2}{c}{\textbf{Strategy Indiv.}} \\
    \cmidrule(lr){5-6} \cmidrule(lr){7-8}
    & & & & (Exact) & (Partial) & (Type A)$^{\dagger}$ & (Type B)$^{\ddagger}$ \\
    \midrule
    Zero-Shot        & 99.01 & 98.19 & 88.16 & 22.58 & 61.07 & 21.05 & 24.11 \\
    Few-Shot         & 99.51 & 99.34 & 81.41 & 19.71 & 60.81 & 18.20 & 21.22 \\
    Chain-of-Thought & 99.18 & 99.34 & 88.76 & 27.24 & 61.32 & 26.80 & 27.68 \\
    \textbf{Ours} (Qwen-3)           & 99.01 & 98.85 & 58.55 & 35.36 & 35.36 & 35.36 & \textbf{54.11} \\ 
    \textbf{Ours} (Llama-3.3)   & \textbf{99.51} & \textbf{99.49} & \textbf{89.47} & \textbf{44.16} & \textbf{96.45} & \textbf{43.10} & 45.22 \\
    \bottomrule
    \specialrule{0em}{2pt}{2pt}
    \multicolumn{8}{l}{\footnotesize $^{\dagger}$ \textbf{Type A}: Funding/Access (e.g., Creates a fund, enables access to a funding stream).} \\
    \multicolumn{8}{l}{\footnotesize $^{\ddagger}$ \textbf{Type B}: Incentives/Permissions (e.g., Creates an incentive for change, expressly allows activity).} \\
  \end{tabular}
  \label{tab:health_acc}
\end{table*}

We next examine performance on the food system classification task, where each policy is annotated with six binary labels. Table \ref{tab:food_results} shows the accuracy of different food categories. Overall, performance varies significantly across categories depending on semantic complexity. Categories such as ``Grow'' and ``Make'', which rely on relatively standardized terminology, achieve high accuracy across all methods. Our framework using Llama-3.3 achieves the highest performance in ``Grow'' (95.39\%) and maintains competitive results in ``Make'' (95.16\%). In contrast, more abstract categories present greater challenges. The ``Get'' category, which involves indirect policy effects on food access, yields the lowest baseline performance, with Zero-Shot achieving only 36.01\%. Our framework improves this substantially to 63.13\%, outperforming Zero-Shot and CoT, and remaining competitive with Few-Shot. Similarly, in ``Process'' and ``Surplus'', our method maintains strong and stable performance, while the ``Distribute'' category remains a consistent bottleneck across all approaches, with accuracies clustered around 70\%. Compared to prompt-based baselines, Qwen-3 demonstrates strong performance in specific categories, achieving the highest accuracy in ``Surplus'' (96.22\%), ``Process'' (95.56\%), ``Make'' (96.05\%), and ``Distribute'' (72.20\%). However, its performance is less balanced, as evidenced by a significant drop in ``Grow'' (65.95\%) and lower performance in ``Get'' (49.34\%). This indicates that strong single-category performance does not necessarily translate into robust multi-label understanding across domains.

\begin{table}[htbp]
\centering
\caption{Accuracy (\%) performance comparison of multi-label food category extraction.}
\label{tab:food_results}
\small
\begin{tabular}{l cccccc}
\toprule
\textbf{Methods} & \textbf{Grow} & \textbf{Surplus} & \textbf{Process} & \textbf{Make} & \textbf{Dist.} & \textbf{Get} \\
\midrule
Zero-Shot        & 94.04 & 93.35 & 80.96 & 88.07 & 71.10 & 36.01 \\
Few-Shot         & 94.49 & 93.16 & 88.25 & 95.16 & 69.35 & \textbf{64.97} \\
Chain-of-Thought & 94.49 & 94.95 & 90.37 & 78.21 & 68.81 & 63.30 \\
\textbf{Ours} (Qwen-3)          & 65.95 & \textbf{96.22} & \textbf{95.56} & \textbf{96.05} & \textbf{72.20} & 49.34 \\
\textbf{Ours} (Llama-3.3)    & \textbf{95.39} & 93.32 & 90.78 & 95.16 & 69.82 & 63.13 \\
\bottomrule
\end{tabular}
\end{table}

For food category extraction, to further evaluate method performance at the policy level, we analyze the distribution of correctly predicted labels for each policy, i.e., multi-label accuracy. Table \ref{tab:multi_label_consistency} reports the percentage of policies for which a method correctly predicts exactly $K$ food categories, where $K$ ranges from 0 to 6. This distribution provides a more detailed view of method performance than aggregate accuracy alone, revealing both the consistency of predictions and the frequency of complete successes or failures. Baseline methods such as Zero-Shot, Few-Shot, and CoT exhibit relatively stable distributions, with most policies concentrated in the range of four to five correct food categories. For example, Zero-Shot peaks at five correct food categories (42.93\%), while Few-Shot and CoT distribute more evenly across four and five correct food categories, indicating moderate but consistent performance. In contrast, our proposed framework achieves notably stronger high-end performance across both model backbones. Our method using Qwen-3 achieves the highest proportion of fully correct predictions among all methods (six food categories correct, 28.78\%) and concentrates the largest share of policies at four correct labels (45.39\%), reflecting a strong ability to precisely identify the relevant food system stages within individual policies. Our method using Llama-3.3 similarly excels in producing fully or near-fully correct predictions, with a combined 51.64\% of policies predicted with five or six correct categories. While its distribution is more bimodal, suggesting the framework either captures multi-label relationships with high fidelity or encounters difficulty with particularly ambiguous policy documents. This characteristic reflects the framework's strength in high-confidence, high-precision extraction rather than a tendency toward uniform but partial predictions. Together, these results demonstrate that the role-based framework consistently pushes performance toward complete label correctness, a key advantage over baselines that plateau at moderate accuracy across the board.

\begin{table*}[htbp]
\centering
\caption{Distribution of correctly predicted food categories per policy across methods. Each column reports the percentage of policies for which exactly $K$ (from 0 to 6) food categories are predicted correctly.}
\label{tab:multi_label_consistency}
\small
\begin{tabular}{l ccccccc}
\toprule
\textbf{Methods} & \textbf{6} & \textbf{5} & \textbf{4} & \textbf{3} & \textbf{2} & \textbf{1} & \textbf{0} \\
\midrule
Zero-Shot        & 14.31 & \textbf{42.93} & 28.78 & 11.84 & 1.32 & 0.82 & 0.00 \\
Few-Shot         & 22.70 & \textbf{31.41} & 29.61 & 12.66 & 2.63 & 0.99 & 0.00 \\
Chain-of-Thought & 23.36 & \textbf{37.34} & 24.51 & 11.84 & 2.63 & 0.33 & 0.00 \\
\textbf{Ours} (Qwen-3)          & 28.78 & 23.19 & \textbf{45.39} & 1.32 & 0.49 & 0.16 & 0.66 \\
\textbf{Ours} (Llama-3.3)    &23.19  & \textbf{28.45} & 12.01 & 6.25 & 1.48 & 0.33 & 28.29 \\
\bottomrule
\end{tabular}
\end{table*}

Overall, these results demonstrate that the proposed role-based framework improves both the completeness of multi-label extraction and per-category accuracy. At the same time, the observed distribution highlights a trade-off between peak performance and prediction stability. This suggests that while structured decomposition enhances the model’s ability to reason over complex policy semantics, further refinement is needed to improve robustness across all instances.

\section{Discussion}
Our results demonstrate that the proposed role-based LLM framework offers meaningful and consistent advantages over standard prompting strategies across nearly every IE task evaluated. While all methods perform comparably on rule-governed fields such as implementation state and effective year where values appear in predictable surface forms, the framework's superiority becomes increasingly pronounced as task complexity grows. This pattern is not incidental: it directly reflects the core design principle of the framework, namely that decomposing complex extraction into specialized, domain-grounded roles enables the kind of structured reasoning that general-purpose prompting strategies cannot reliably produce.

The most striking gains are observed in legal strategy extraction, the most semantically demanding subtask in our evaluation. Baseline methods plateau between 19-27\% exact-match accuracy, reflecting their fundamental inability to handle multi-label classification when category boundaries are abstract, overlapping, and not lexically grounded in the source text. Our framework achieves 44.16\% exact-match accuracy with Llama-3.3, nearly double the strongest baseline, and 96.45\% under partial-match evaluation, far exceeding the approximately 61\% achieved by zero-shot and CoT approaches. This result is particularly significant because legal strategy recognition requires the model to simultaneously infer policy intent, map actions to institutional mechanisms, and assign multiple applicable strategies from a structured typology. By equipping the Legal Strategy Specialist role with explicit category definitions, edge case handling, and step-by-step reasoning requirements, the framework transforms an under-constrained classification problem into a tractable, guided inference task. No baseline method, regardless of prompting sophistication, replicates this effect, because none provides the role-specific domain scaffolding that makes such reasoning possible.

Our framework's advantages extend equally to food system classification. Across the six binary food categories, our Llama-3.3 configuration achievea the highest or near-highest accuracy in five of six categories, including 95.39\% on ``Grow" and 95.16\% on ``Make". More importantly, it delivers these results with greater consistency across categories than any baseline, reflecting the Food System Expert role's use of precise, per-category definitions that reduce inter-label ambiguity. The ``Get" category, which represents the hardest classification target in this subtask due to its broad and polysemous nature, illustrates our framework's impact most clearly: zero-shot performance falls to just 36.01\%, while our framework raises this to 63.13\%, an improvement of over 27\% points. This gain reflects the role-specific prompt's ability to provide a restrictive, contextually grounded definition of ``Get" that constrains the model's interpretation in ways that general prompting cannot.

The multi-label distribution analysis in Table \ref{tab:multi_label_consistency} reinforces these findings at the policy level. While baseline methods tend to produce moderate, uniform performance peaking around four to five correct food categories per policy, our framework using Llama-3.3 achieves the strongest concentration of fully or near-fully correct predictions, with 51.64\% of policies predicted with five or six correct categories. This characteristic bimodal distribution is itself informative: it indicates that the framework either captures multi-label relationships with high fidelity or encounters genuine difficulty with particularly ambiguous documents. Such a pattern is consistent with principled, high-confidence reasoning where the model commits strongly when evidence is clear rather than the diffuse, uniformly moderate performance often observed in purely prompt-based approaches. Our Qwen-3 configuration further validates the framework's generalizability, achieving the highest rate of fully correct six-category predictions (28.78\%) and leading performance in ``Surplus", ``Process", and ``Make", demonstrating that the role-based design yields gains independent of the underlying model backbone.

The reasons why competing methods fall short are instructive precisely because they highlight what the framework uniquely provides. Zero-shot prompting lacks structured domain guidance entirely, leaving the model to draw on generic representations of policy language that are inconsistent for low-frequency or abstract labels. CoT prompting introduces step-by-step reasoning but without role-specific framing, producing marginally better results on some tasks while still failing to constrain the model's interpretation of semantically complex categories. Few-shot prompting, despite providing labeled examples, achieves the lowest legal strategy accuracy of all baselines (19.71\%), suggesting that example-based guidance alone without structured definitional scaffolding may actively introduce noise in complex multi-label tasks. Our framework addresses all of these failure modes simultaneously: it grounds reasoning in explicit domain knowledge, decomposes the extraction problem into cognitively tractable subtasks, and enforces structured output formats that reduce hallucination and label drift.

Several remaining challenges point toward directions in which the framework's already strong performance can be further strengthened. Legal strategy extraction, despite substantial improvement over baselines, remains difficult in cases where multiple strategies apply conjunctively and where the relevant policy intent is distributed across multiple paragraphs rather than concentrated in a single span. Integrating retrieval-augmented generation (RAG) \cite{xu2025mega}, providing the LLM Legal Strategy Specialist with curated definitional documents or canonical labeled examples at inference time, could reduce disambiguation errors for conceptually adjacent categories, such as distinguishing ``creates an incentive" from ``expressly allows something". Similarly, incorporating uncertainty-aware abstention mechanisms \cite{zhang2026alarm} or quality control process \cite{zhang2026team} would filter confident yet unsupported attribute values when relevant evidence is absent from the source text. Such safeguards enhance the reliability and evidentiary grounding of the framework’s outputs, thereby strengthening their suitability for downstream policy analysis and decision-making.

These directions notwithstanding, the core contribution of this work is clear: structured role decomposition with domain-grounded prompting is a practically effective and generalizable solution to the challenges of policy IE. By bridging the gap between lightweight general-purpose prompting and resource-intensive fine-tuning or agent-based systems, the framework offers a scalable and transparent methodology that outperforms existing alternatives on the tasks that matter most, precisely those where policy language is most complex, most ambiguous, and most consequential for downstream decision-making.

\section{Conclusion}
The increasing volume and complexity of public health policy documents demand scalable, reliable methods for automated IE. While LLMs offer considerable promise in this area, their direct application to health policy data remains fraught with hallucinations, misclassifications, and omissions that undermine analytical trustworthiness. This work addresses these limitations through a role-based prompting framework that assigns three domain-grounded LLM roles, i.e., Policy Analyst, Legal Strategy Specialist, and Food System Expert, each guided by structured classification criteria and task-specific reasoning requirements designed to mirror expert analytical workflows. Evaluated against 608 healthy food policies from the HFPP database, our framework consistently outperforms Zero-Shot, Few-Shot, and CoT baselines across the most semantically demanding extraction tasks. Policy type classification reaches 89.47\% accuracy, while legal strategy extraction achieves 44.16\% exact-match accuracy, nearly double the best-performing baseline, and 96.45\% under partial-match evaluation. Across six food system categories, the framework delivers competitive or superior performance, with particularly strong results in ``Grow" (95.39\%) and ``Make" (95.16\%). Critically, these gains are not model-specific: comparable improvements observed with Qwen-3 confirm that the role-based design contributes independently of the underlying model backbone. Beyond raw performance, our framework advances a principled middle ground between lightweight general-purpose prompting and resource-intensive fine-tuning, offering a transparent and reproducible methodology well-suited to rapidly evolving policy domains. Its practical implications extend to more efficient evidence-based policymaking and greater accountability in public health governance. Future work should explore extending this framework to other policy domains, integrating RAG to sharpen disambiguation in complex multi-label tasks, and developing uncertainty-aware mechanisms that allow the model to abstain when source evidence is insufficient, further strengthening the reliability of LLM-assisted policy analysis at scale.

\section*{Acknowledgments}
The authors thank Dr. Juhua Hu and Zijing Wei from University of Washington for their valuable feedback and assistance. The authors also acknowledge the Healthy Food Policy Project for generously sharing data that supported this research.

\bibliographystyle{unsrtnat}
\bibliography{sample-base}

\end{document}